\documentclass[runningheads]{llncs}

\usepackage{eccv}

\usepackage{eccvabbrv}

\usepackage{graphicx}
\usepackage{booktabs}
\usepackage{multirow}
\usepackage{tabularx}
\usepackage{subcaption} 
\usepackage{float}
\usepackage[accsupp]{axessibility}  

\usepackage{listings}
\usepackage{xcolor}
\usepackage{tcolorbox} 
\lstdefinestyle{PythonStyle}{
    language=Python,
    basicstyle=\ttfamily\fontsize{6.6}{7.8}\selectfont, 
    backgroundcolor=\color{gray!6}, 
    keywordstyle=\color{blue}, 
    commentstyle=\color{green!55!black}, 
    stringstyle=\color{orange}, 
    numbers=left, 
    numberstyle=\tiny\color{gray}, 
    breaklines=true, 
    frame=single, 
    framexleftmargin=5mm, 
    captionpos=b, 
    showstringspaces=false,
}
\newtcolorbox[blend into=figures]{codebox}[1][]{
    enhanced,
    arc=0pt,
    boxrule=0.6pt,
    colback=gray!10,
    colframe=black,
    width=\textwidth,
    sharp corners,
    boxsep=2pt,
    left=3pt,
    #1
}

\usepackage{hyperref}
\usepackage{orcidlink}

\begin{document}

\title{NAMER: Non-Autoregressive Modeling for Handwritten Mathematical Expression Recognition} 

\titlerunning{NAMER: Non-Autoregressive Modeling for HMER}

\author{Chenyu Liu\inst{1,2} \and
Jia Pan\inst{2} \and
Jinshui Hu\inst{2} \and
Baocai Yin\inst{2} \and 
Bing Yin\inst{2}\and \\ 
Mingjun Chen\inst{2}\and 
Cong Liu\inst{2}\and 
Jun Du\inst{1}$^\dag$ \and 
Qingfeng Liu\inst{1,2}
}
\renewcommand{\thefootnote}{}
\footnotetext[2]{$^\dag$ Corresponding author.}

\authorrunning{C. Liu et al.}

\institute{University of Science and Technology of China, Hefei, China \and
iFLYTEK Research, Hefei, China \\
\email{cyliu7@mail.ustc.edu.cn, jundu@ustc.edu.cn}\\
}
\maketitle

\begin{abstract}
Recently, Handwritten Mathematical Expression Recognition (HMER) has gained considerable attention in pattern recognition for its diverse applications in document understanding. Current methods typically approach HMER as an image-to-sequence generation task within an autoregressive (AR) encoder-decoder framework. However, these approaches suffer from several drawbacks: 1) a lack of overall language context, limiting information utilization beyond the current decoding step; 2) error accumulation during AR decoding; and 3) slow decoding speed. To tackle these problems, this paper makes the first attempt to build a novel bottom-up \textbf{N}on-\textbf{A}utoRegressive \textbf{M}odeling approach for H\textbf{MER}, called NAMER. NAMER comprises a Visual Aware Tokenizer (VAT) and a Parallel Graph Decoder (PGD). Initially, the VAT tokenizes visible symbols and local relations at a coarse level. Subsequently, the PGD refines all tokens and establishes connectivities in parallel, leveraging comprehensive visual and linguistic contexts. Experiments on CROHME 2014/2016/2019 and HME100K datasets demonstrate that NAMER not only outperforms the current state-of-the-art (SOTA) methods on ExpRate by 1.93\%/2.35\%/1.49\%/0.62\%, but also achieves significant speedups of 13.7$\times$ and 6.7$\times$ faster in decoding time and overall FPS, proving the effectiveness and efficiency of NAMER.

\end{abstract}

\section{Introduction}
\label{sec:intro}

Handwritten Mathematical Expression Recognition is a fundamental task in the OCR and pattern recognition field, playing a wide-ranging role in document understanding, teaching and education, as well as office automation. With the developments of deep learning \cite{alexnet, resnet, densenet, transformer}, sequence modeling \cite{NMT, LAS, Coverage}, and text recognition \cite{crnn, aster, str-survey} over the past decade, more and more HMER algorithms \cite{dwap, treedecoder, bttr, abm, HME100K, CAN-ECCV, comer} have been proposed with large performance improvements.
\begin{figure}[t]
  \centering
  \small
  \begin{subfigure}{0.8\linewidth}
  	\centering
    \includegraphics[width=1.0\linewidth]{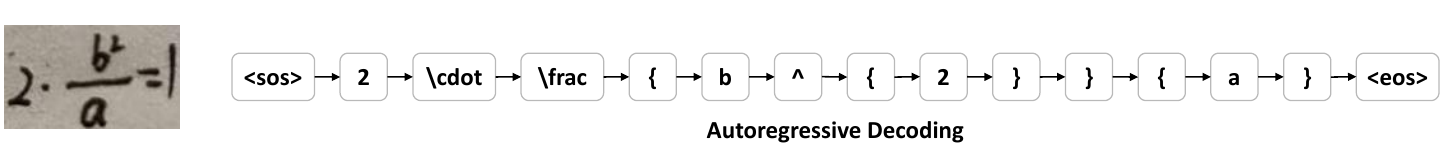}
    \caption{Existing AR modeling of HMER.}
    \label{fig:pipeline-a}
  \end{subfigure}
 
   \begin{subfigure}{0.8\linewidth}
   \centering
    \includegraphics[width=1.0\linewidth]{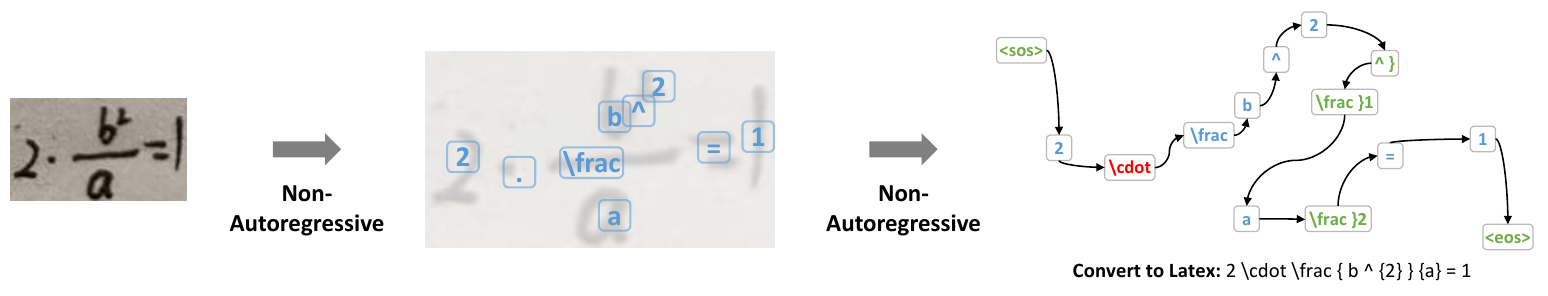}
    \caption{A brief illustration of NAMER.}
    \label{fig:pipeline-b}
  \end{subfigure} 
  \caption{Comparison between existing AR based HMER methods and the proposed NAMER. All processes in NAMER follow a NAR manner.}
  \label{fig:pipeline}
\end{figure}
Different from text recognition and speech recognition,  there are many complicated symbol relations in mathematical expressions, making HMER more challenging. For example, a mathematical expression ``\textbackslash frac\{x\textasciicircum\{y\}\}\{z\textunderscore\{1\}\}'' has three relations: fraction, exponent, and subscript. To tackle this, researchers have proposed to convert HMER to an image-to-markup \cite{markup} generation task. Hence, lots of encoder-decoder based sequence modeling methods \cite{NMT, LAS, Coverage} have been improved to deal with HMER successfully. Specifically, given an image, an encoder is firstly used to extract visual features, and an attention based decoder is then designed to recognize LaTeX strings token by token, following an AR manner. See \cref{fig:pipeline-a} for a brief illustration. Moreover, researchers have persistently strived to advance HMER, focusing on refining both structured modeling and contextual inference. Tree based \cite{treedecoder, TSDNet, tdv2, cheng2024} and syntax-aware \cite{HME100K} targets are proposed for image-to-sequence prediction with robuster structural recognitions, while symbol counting \cite{CAN-ECCV}, bidirectional modeling \cite{bttr, abm} and coverage modeling \cite{comer} are designed to enhance the contextual inference abilities of HMER.

Despite great performance improvements on HMER, there are still some weaknesses of current methods. Firstly, current models lack the ability to flexibly utilize the overall visual or linguistic context features. Although the bidirectional and coverage modeling designs have already improved the utilization of context, current models are still prone to misclassify ambiguous symbols that human can easily recognize them correctly by using vision-language context informations. Secondly, because of the mechanism that current models need to recognize tokens in an autoregressive manner, inference speeds of these models are slow.

To jointly improve the utilization of overall vision-language infos and recognition efficiency in HMER, we propose a new Non-Autoregressive Modeling framework for HMER, called NAMER. Our motivation is from the recognition ability and process of humans: (1) humans don't recognize and understand HMEs in a predefined LaTeX string or symbol tree order, instead, humans usually start from recognizing symbols in an approximate left-to-right order and determining relations between these symbols; (2) humans have the strong ability to recognize from arbitrary local symbols to global results in a bottom-up manner, and when facing ambiguous symbols, the overall bottom information of local visual features and local recognition results are dynamically used. Inspired by this, we aim to build a HMER system which have two key characteristics: one is the mechanism of recognizing or tokenizing local symbols in parallel, the other is recovering the connectivities and relations between these tokens in a non-autoregressive (NAR) manner. To achieve this, NAMER adopts a two-stage, bottom-up approach while keeping end-to-end training. As illustrated in \cref{fig:pipeline-b}, the first stage predicts all visible symbols and local relation tokens, including some imaginary relationship symbols like ``\textasciicircum'', along with their approximate locations in parallel, providing coarse symbol-instance level results. The second stage refines these predicted coarse tokens while simultaneously predicting the connectivity probabilities among them in parallel. Both stages follow a NAR manner, and the resultant output constructs a Directed Acyclic Graph (DAG) that can easily be converted to the LaTeX format or Symbol Layout Trees\cite{treedecoder}. Consequently, NAMER is a NAR recognition system that doesn't require decoding tokens sequentially, resulting in significant speedups. Moreover, the two-stage NAR design not only enhances flexibility in revising misclassified tokens by leveraging global visual and linguistic context features more effectively but also ensures robustness in reconstructing the structure of HMEs from local to global.

To verify the effectiveness of NAMER, we conducted comprehensive experiments on the CROHME 2014 \cite{crohme14}, 2016 \cite{crohme16}, 2019 \cite{crohme19} datasets and HME100K \cite{HME100K} dataset. In short, NAMER significantly enhances decoding speed while maintaining SOTA performance. Across the CROHME datasets, NAMER achieves ExpRates of 60.51\%, 60.24\%, and 61.72\% without augmentations on CROHME 2014, 2016, and 2019, respectively, surpassing SOTA methods (58.57\%, 57.89\%, 59.71\%) by an average of 1.93\%. Regarding inference speed, NAMER demonstrates a 13.7$\times$ and 6.70x$\times$ improvement in decoding time and overall inference time, respectively, compared to the previous SOTA method, CoMER \cite{comer}. In summary, the main contributions of this paper are as follows:
\begin{enumerate}

  \item NAMER represents the first endeavor, to the best of our knowledge, in constructing a non-autoregressive HMER system without extra annotations;
  \item The VAT, PGD modules, and Bipartite Matching training strategy are meticulously designed to facilitate a bottom-up NAR decoding mechanism;
  \item NAMER demonstrates significant improvements in both decoding speed and memory usage while achieving state-of-the-art performance.

\end{enumerate}

\section{Related Work}
\label{sec:related_work}

\subsection{HMER Methods}
Traditional HMER methods consist of three parts: symbol segmentation, classification and structure analysis. These methods firstly segment inputs into independent symbols, and then classify them using HMM \cite{hmm1, hmm2, hmm3}, Elastic Matching \cite{Elastic}, or Support Vector Machines \cite{svm}. As for structure analysis, formal grammars are carefully designed to model the 2D and syntactic structures of formulas, where \cite{graphgrammar} proposes graph grammar to recognize mathematical expression and \cite{errorhmer} designs correction mechanism into a parser based on definite clause grammar. However, limited by explicit segmentation and manual rules, performance of traditional methods is still far from real applications.

Recently, deep learning has made huge success in HMER. Deng \etal \cite{markup} firstly showed the ability of image-to-markup generation using neural networks, and Zhang \etal also proposed WAP \cite{wap} and DWAP \cite{dwap} that apply an attention based encoder-decoder to generate LaTeX strings for HMER in an AR manner. After DWAP, AR-based encoder-decoders with image-to-markups become the fundamental approach of HMER, and further developments can be divided into two directions: decoding target design and decoding architecture design. 

In terms of decoding target, Zhang \etal proposed TreeDecoder \cite{treedecoder} to use tree structure sequence as targets instead of LaTeX strings. Wu \etal futher simplify TreeDecoder and improve generalization capability as TD-V2 \cite{tdv2}, while SSD \cite{ssd} propose a new breadth-first markup named structural string that enhance both language modeling and hierarchical structural modeling. In addition, TSDNet \cite{TSDNet} design a transformer-based tree decoder that can better capture complicated correlations. More recently, Yuan \etal proposed Syntax-Aware Network (SAN), in which syntactic information is effectively merged into the AR decoder while maintaining efficiency. The improvement of decoding target design allow HMER models to simultaneously focus on both HME structure and local symbols, bringing a robuster recognition ability.

In terms of decoding architecture, PAL \cite{pal} and PAL-V2 \cite{palv2} design a discriminator along with an adversarial learning method to improve the robustness to writing styles. BTTR \cite{bttr} firstly use transformer decoders for HMER and put forward a bi-directionally training framework to enhance language modeling in HMER, then ABM \cite{abm} improve it with bi-directionally mutual learning and multi-scale coverage attention. Recently, CAN \cite{CAN-ECCV} designs extra heads for symbol counting, which helps correct the prediction errors of original decoders. Moreover, CoMER \cite{comer} proposed general attention refinement module and different coverage attention fusion modules to better model attentions in transformer-based HMER decoders, outperforming existing methods by a large margin.

\subsection{NAR for HMER}
At present, there are no NAR-based HMER methods available. Here, we conduct a survey of NAR sequence modeling works in other areas.

The most common algorithm for NAR sequence recognition is CTC \cite{CTC}, which is widely used in scene text recognition \cite{crnn, 2dctc} and speech recognition \cite{CTC,speech-ctc}. However, due to the 2D layouts and relations in HMEs, there is no efficient way to adopt CTC or 2D-CTC \cite{2dctc} in HMER currently.

Another approach involves NAR transformers, previously explored in neural machine translation \cite{nanmt} and scene text recognition \cite{parseq, pimnet, ipad}. In \cite{nanmt}, a direct implementation of the NAR transformer decoder enables the simultaneous translation of all tokens, whereas PARSeq \cite{parseq} adopts a query-based parallel decoding paradigm, and \cite{pimnet, ipad} improve upon this paradigm with iterative refinements. Nevertheless, due to the structural layout and the presence of imaginary tokens in HME, these NAR transformer-based methods are unable to handle HMER. 

The most related work to NAMER is Graph-to-Graph (G2G) \cite{g2g}, which is designed for online HMER. G2G firstly treat each stroke trajectory as a input graph node, then use graph neural networks to encode graph features and decode Symbol Label Tree (SLT) based graph in an AR mode. However, G2G need trajectory-symbol level extra annotations for training and online sequential trajectory positions for testing, hence G2G cannot be applied to general HMER.
\label{method}

\section{Methodology}

\subsection{Overview}
 \begin{figure*}[t!]
    \centering
    \includegraphics[width=0.91\linewidth]{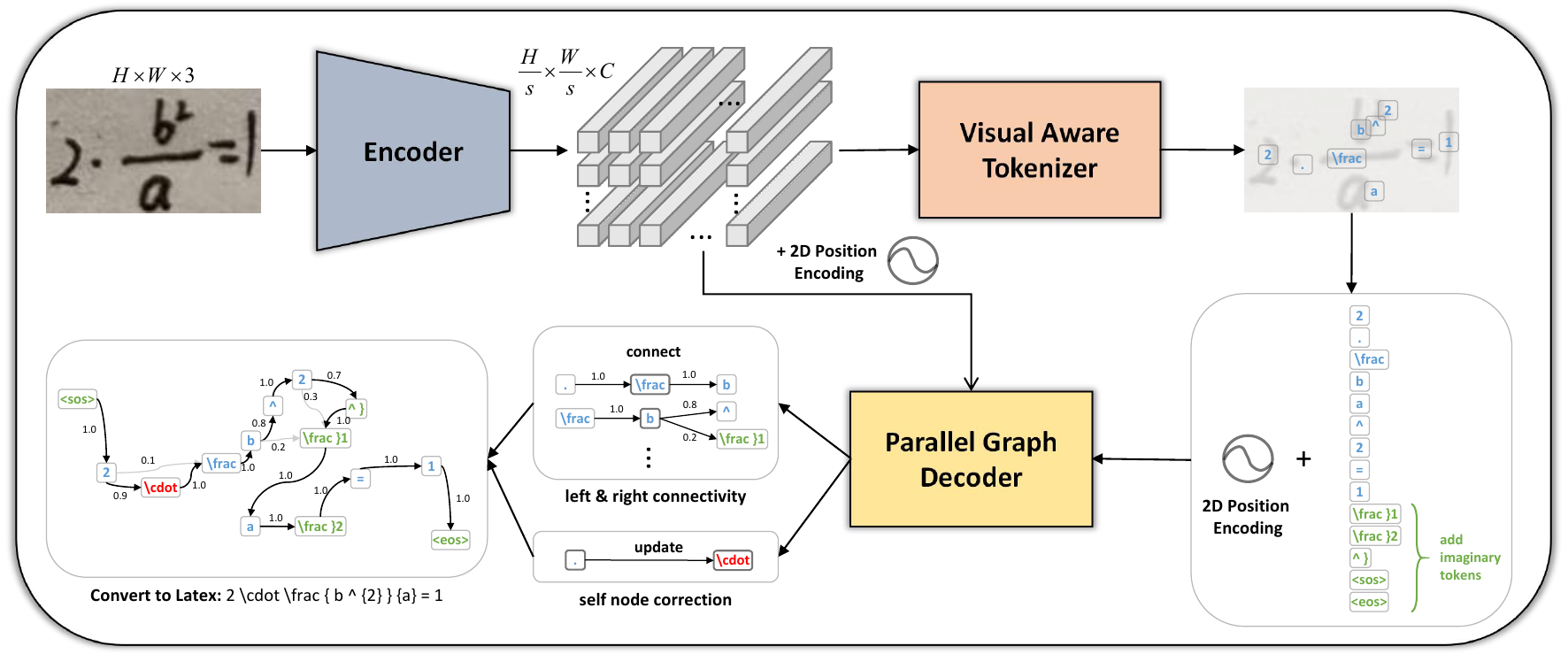}
    \caption{Overall framework of the proposed method. NAMER consists of an encoder, a visual aware tokenizer, and a parallel graph decoder. For an input image, VAT firstly predicts all visible symbols and local relations at a coarse level, then PGD will revise these tokens and establish the connectivity between them. After PGD, a DAG is used for converting results to other format like LaTeX. All modules are in NAR manners.}
    \label{fig:framework}
  \end{figure*}
The overall framework of NAMER is illustrated in \cref{fig:framework}. Given an image $\mathbf{I}\in \mathbb{R}^{3\times H \times W}$ and its LaTeX label $\mathbf{y} = \{y_1, y_2, \ldots, y_L\}$, a backbone encoder first encodes it into a downsampled visual feature $\mathbf{F}\in \mathbb{R}^{C\times \frac{H}{s} \times \frac{W}{s} }$. Subsequently, the VAT module predicts all visible symbols and local relation tokens along with their locations in parallel. After adding the corresponding imaginary tokens (e.g., an ending token ``\}'' for ``\textasciicircum''), initial coarse results are constructed without considering their relations. Refer to the bottom right of \cref{fig:framework} for an illustration.

Next, these initial tokens, along with their 2D position embeddings, are fed into the PGD module. PGD comprises two main branches: one for revising input tokens and the other for predicting the directional connectivity between each token. Both branches are designed in a non-autoregressive manner.

In the final step, leveraging the node correction results obtained in the PGD module and the directional connection scores assigned to each edge, a directed acyclic graph (DAG) is meticulously built. This graph serves as the foundation for tracing a path starting from the node ``\textless sos\textgreater'' and concluding at the node ``\textless eos\textgreater'', thereby yielding the ultimate recognition result.

\subsection{Visual Aware Tokenizer}

\begin{figure}[ht]
  \centering
  \begin{subfigure}{0.47\linewidth}
  \centering
    \includegraphics[width=0.95\linewidth]{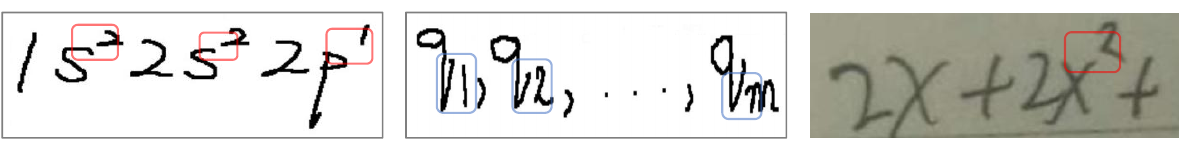}
    \caption{}
    \label{fig:VAT-a}
  \end{subfigure}
  \hfill
  \begin{subfigure}{0.47\linewidth}
  \centering
    \includegraphics[width=0.95\linewidth]{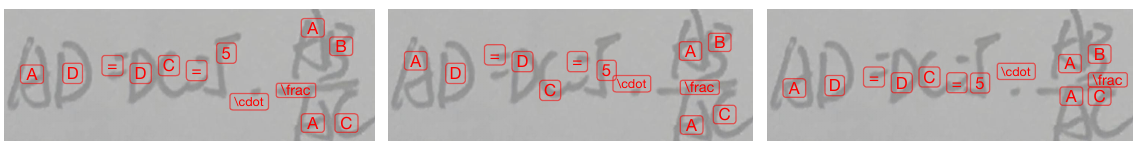}
    \caption{}
    \label{fig:VAT-b}
  \end{subfigure}
    \label{fig:VAT_exp}
  \caption{Feasibility analysis of VAT. (a) Visual patterns of local relation tokens. ``\textasciicircum'' is in red color rect and ``\textunderscore'' is in blue. (b) Coarse visual tokens on an HME image. Though all locations are imprecise, the three cases can all lead to a correct recognition finally.}
\end{figure}

Given a 2D spatial feature map, the VAT module is designed to tokenize all visible elements, including visible symbols and local relation tokens. Visible symbols refer to handwritten characters, while local relation tokens encompass handwritten structural elements (e.g., ``\textbackslash frac'', ``\textbackslash sqrt'') and imaginary relationship symbols (e.g., ``\textunderscore'', ``\textasciicircum''). As depicted in \cref{fig:VAT-a}, similar to other handwritten characters, the ``\textasciicircum'' and ``\textunderscore'' tokens exhibit distinct visual patterns, with relationships in the upper or lower right corner. Therefore, these local relation tokens together with visible symbols are all feasible to predict, and VAT is trained to address them. 
Mathematically, given encoded feature maps $\mathbf{F}_{8\mathrm{x}}\in \mathbb{R}^{C_{1}\times \frac{H}{8} \times \frac{W}{8}}$ and $\mathbf{F}_{16\mathrm{x}}\in \mathbb{R}^{C_{2}\times \frac{H}{16} \times \frac{W}{16}}$, we firstly use two CNN layers to merge an 8x downsampled feature map. This step is taken because a higher resolution feature can effectively capture small symbols:
\begin{equation}
 \mathbf{F}_{8\mathrm{x}}^{\prime} = \mathrm{Conv_{1x1}}({\mathrm{Upsample_{2x}}(\mathbf{F}_{16\mathrm{x}})}),~ \mathbf{F}_{out} = \mathrm{Conv_{3x3}}(\mathrm{Concat}(\mathbf{F}_{8\mathrm{x}}, \mathbf{F}_{8\mathrm{x}}^{\prime}))
  \label{eq:fpn_simple}
\end{equation}
Here, $\mathrm{Conv_{ixi}}(\cdot)$ denotes a standard $i\times i$ Conv-BN-Relu block \cite{resnet}, $\mathrm{Upsample_{2x}(\cdot)}$ denotes a bilinear 2x upsample layer, and $\mathrm{Concat(\cdot)}$ denotes a feature concatenation operation. Then, after a classification head, we can obtain the predicted probabilities at each location of $\frac{H}{8} \times \frac{W}{8}$:
\begin{equation}
 \mathbf{P} = \mathrm{Softmax}(\mathrm{Conv_{3x3}}(\mathbf{F}_{out}))
 \label{eq:cls_head}
\end{equation}
where $\mathbf{P}\in \mathbb{R}^{(K+1)\times\frac{H}{8} \times \frac{W}{8}}$, the number $K$ stands for the total number of all visible elements and the class id of $K+1$ denotes none token $\varnothing$.

\begin{figure}[t]
  \centering
    \includegraphics[width=0.82\linewidth]{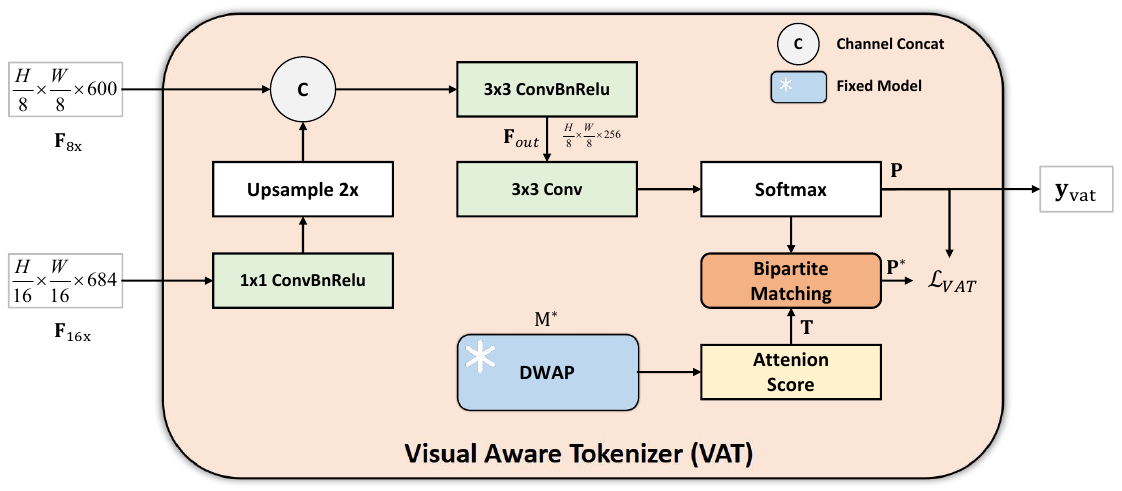}
  \caption{Detailed structure of the proposed VAT module.}
  \label{fig:VAT-framwork}
\end{figure}

For the HMER task, the aforementioned visual tokenization process does not require predicting precise boxes or masks for each token; instead, tokenization categories and approximate positions are usually sufficient to recover the structure of mathematical expressions. As illustrated in \cref{fig:VAT-b}, even if the tokenized positions differ, as long as the symbol instances are distinguishable and their positions are roughly correct, it does not affect subsequent structural recovery. Since character-level annotations are hard to obtain, we propose extracting this information from a fixed HMER model $\mathrm{M^*}$, and a pretrained DWAP \cite{dwap} is used as $\mathrm{M^*}$. However, most HMER models suffer from the attention shift problem, making it impractical to directly obtain symbols' positions from their attention maps. Therefore, we design a bipartite matching-based \cite{bmt, detr} dynamic assignment approach to guide VAT training (see \cref{fig:VAT-framwork}), finding it useful and necessary:
\begin{enumerate}
  \item Forward $\mathrm{M^*}\mathbf{(I, y)}$ under teacher forcing mode and obtain the attention scores $\mathbf{A} \in \mathbb{R}^{L \times \frac{H}{8} \times \frac{W}{8}}$ for the decoding step of $\{y_1, y_2, \ldots, y_L\}$;
  \item Utilize the highest attention scores of each $y_i$ to estimate its spatial position $\mathbf{T} \in \mathbb{Z}^{L \times 2}$, where $\mathbf{T}_l = \mathrm{argmax}(\mathbf{A}_l)$; 
  \item Conduct a bipartite matching process between the prediction set $\mathbf{P}$ and the estimate set $\mathbf{T}$. The cost matrix $\mathbf{D}\in \mathbb{R}^{L\times HW}$ is designed to assign the highest predicted score of label $y_i$ within a $k_{m} \times k_{m}$ kernel to each estimated $y_i$. Utilizing the Hungarian algorithm for optimal assignment, these results are converted to the final training target $\mathbf{P}^* \in \mathbb{Z}^{\frac{H}{8} \times \frac{W}{8}}$. Refer to the Supplementary Materials for matching details.
\end{enumerate}
After assigning targets, the loss function of VAT can be written as:
\begin{equation}
  \mathcal{L}_{\mathrm{VAT}}=\mathrm{CrossEntropy}(\mathbf{P}, \mathbf{P}^{*})
\end{equation}
During testing, VAT simply choose symbols that have highest classification scores on each position, and filter the none token $\varnothing$ out:
\begin{equation}
  \mathbf{y}_{\mathrm{vat}}=\{\mathrm{y_i}|\mathrm{y_i} \in \arg\max \mathbf{P} \cap \mathrm{y_i} \neq \varnothing\}
\end{equation}
\subsection{Parallel Graph Decoder}
\label{sec:PGD}
\begin{figure}[t]
  \centering
    \includegraphics[width=0.82\linewidth]{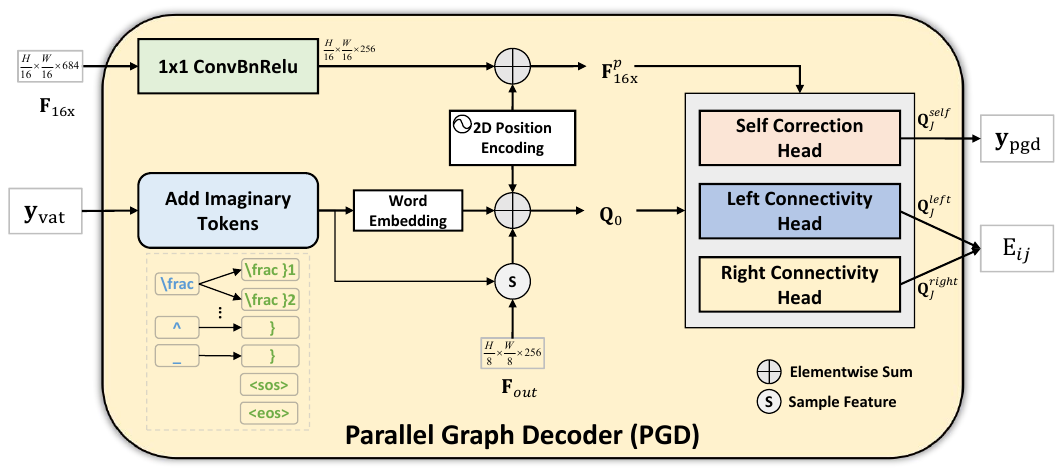}
  \caption{Detailed structure of the proposed PGD module.}
  \label{fig:PGD-framwork}
\end{figure}

The detailed structure of PGD is shown in \cref{fig:PGD-framwork}. After the VAT module, we obtain all visible tokens $\mathbf{y}_{\mathrm{vat}}$. Then, for the structural symbols in $\mathbf{y}_{\mathrm{vat}}$, we add their respective attached imaginary tokens, updating $\mathbf{y}_{\mathrm{vat}}$ as $\mathbf{y}_{\mathrm{vat}}^{\prime}$. The imaginary symbols for each structural symbol are different. For example, ``\textasciicircum'' has one ``\}'' ending imaginary token, while ``\textbackslash frac'' has two ending imaginary tokens for both numerator and denominator (see Supplementary Materials for details). 

PGD consists of three parallel heads: Self-node Correction Head (SCH), Left Connectivity prediction Head (LCH), and Right Connectivity prediction Head (RCH). Each head has the same architecture but different training targets. All heads adopt a transformer-based non-autoregressive decoder with cross-attention in it. Mathematically, for each layer $j$:
\begin{equation}
\label{qkv}
\centering
\begin{aligned}
	& \mathbf{Q}_{j}^{\prime} = \mathbf{Q}_{j} + \mathrm{Attention}(\mathbf{Q}_{j}, \mathbf{F}_{16\mathrm{x}}^{p}, \mathbf{F}_{16\mathrm{x}}^{p}), \\
	& \mathbf{Q}_{j+1} = \mathrm{Transformer}(\mathbf{Q}_{j}^{\prime})
\end{aligned}
\end{equation}
where $\mathbf{F}_{16\mathrm{x}}^{p}$ stands for $\mathbf{F}_{16\mathrm{x}}$ with 2d position encoding, $\mathrm{Attention}(\mathrm{query}, \mathrm{key}, \mathrm{value})$ refers to a standard attention layer and $\mathrm{Transformer}(\cdot)$ stands for a simple transformer layer \cite{transformer}. $\mathbf{Q}_{j}$ denotes the query input for every layer $j$, and $\mathbf{Q}_{0}$ is initialized as the sum of VAT tokens' visual feature, position encoding and word embedding, as depicted in \cref{fig:PGD-framwork}.

After parallel decoding across three heads, we obtain the final features of each head: $\mathbf{Q}_{J}^{self}$, $\mathbf{Q}_{J}^{left}$, and $\mathbf{Q}_{J}^{right}$.

The goal of self-node correction head is to learn an implicit vision-language model to correct misclassified nodes, using the inputs of the visual features, the predicted language tokens,  and their approximate locations. The correction process includes revising a symbol or deleting a symbol. So the training of $\mathbf{Q}_{J}^{self}$ can still convert to a classification problem:
\begin{equation}
\begin{aligned}
\mathbf{p}_{y}^{self} = \mathrm{Softmax}(\mathbf{W}_{self}^{T}\cdot\mathbf{Q}_{J}^{self}&), ~ \mathbf{y}_{\mathrm{pgd}}=\arg\max (\mathbf{p}_{y}^{self}), \\ 
\mathcal{L}_{\mathrm{PGD}}^{\mathrm{self}} = \mathrm{CE}(\mathbf{p}&_{y}^{self},\mathbf{y}_\mathrm{assign}^{\mathrm{self}})
  \end{aligned}
\end{equation}
here $\mathbf{W}_{self}^{T}$ is a trainable FC layer, $\mathbf{y}_\mathrm{assign}^{\mathrm{self}}$ is the node's GT target according to the VAT's bipartite matching results.

Moreover, the goal of the two connectivity heads are to learn an explicit layout model in parallel to dynamically choose the left and right token for each node. This task is straightforward: for each token in VAT's predicted $\mathbf{y}_{\mathrm{vat}}^{\prime}$, choose another token from $\mathbf{y}_{\mathrm{vat}}^{\prime}$ as its left side or right side token, so that PGD can adopt this connectivity as an edge and build a overall directed graph. The selection processes are modeled as attentions from the initial query $\mathbf{Q}_{0}$ over the computed layout features:
\begin{equation}
\begin{aligned}
	\mathcal{L}_{\mathrm{PGD}}^{\mathrm{left}} &= \mathrm{CE}(\mathrm{Softmax}(\mathbf{Q}_{0}^{T}\cdot\mathbf{Q}_{J}^{left}), \mathbf{y}_\mathrm{assign}^{\mathrm{left}}), \\
	\mathcal{L}_{\mathrm{PGD}}^{\mathrm{right}} &= \mathrm{CE}(\mathrm{Softmax}(\mathbf{Q}_{0}^{T}\cdot\mathbf{Q}_{J}^{right}), \mathbf{y}_\mathrm{assign}^{\mathrm{right}})
\end{aligned}
\end{equation}

Similar to $\mathbf{y}_\mathrm{assign}^{\mathrm{self}}$, once the bipartite matching in VAT is done, all nodes' GT connectivity target $\mathbf{y}_\mathrm{assign}^{\mathrm{left}}$, $\mathbf{y}_\mathrm{assign}^{\mathrm{right}}$ can be obtained. The loss of whole PGD is:
\begin{equation}
  \mathcal{L}_{\mathrm{PGD}} = \mathcal{L}_{\mathrm{PGD}}^{\mathrm{self}} + \mathcal{L}_{\mathrm{PGD}}^{\mathrm{left}} + \mathcal{L}_{\mathrm{PGD}}^{\mathrm{right}}
\end{equation}

And the overall training loss of NAMER is:
\begin{equation}
  \mathcal{L}_{all} = \mathcal{L}_{\mathrm{VAT}} + \lambda\cdot\mathcal{L}_{\mathrm{PGD}}
 \label{eq:loss_all}
\end{equation}
where $\lambda$ is a hyper-parameter for balancing the training of VAT and PGD.

\subsection{Path Selection}
After PGD, we can obtain all revised nodes $\mathrm{V}=\arg\max(\mathbf{p}_{y}^{self})$, and connectivity scores from node $\mathrm{V}_i$ to node $\mathrm{V}_j$:
\begin{equation}
  \mathrm{E}_{ij} = \mathrm{Softmax}(\mathbf{Q}_{0}(j)^{T}\cdot\mathbf{Q}_{J}^{left})_i + \mathrm{Softmax}(\mathbf{Q}_{0}(i)^{T}\cdot\mathbf{Q}_{J}^{right})_j
\end{equation}
 The directed edge score $\mathrm{E}_{ij}$ is the sum of the right connectivity score from node $\mathrm{V}_i$ to node $\mathrm{V}_j$ and the left connectivity score from node $\mathrm{V}_j$ to node $\mathrm{V}_i$. To minimize unnecessary computations, we delete edges with scores below a hyper-parameter $\epsilon$ (if it fails to partition the graph into two disconnected subgraphs).

So far, we have built a directed acyclic graph $\textless\mathrm{V}, \mathrm{E}\textgreater$ in parallel, with all positive edge weights. Our path selection strategy is to find the longest path from the node \textless sos\textgreater  to the node \textless eos\textgreater. Finding the longest path in a DAG \cite{bellman, dijkstra1959note} with positive edges can be done using a dynamic programming approach based on a topological ordering of the vertices, under the time complexity of $O(V+E)$. It's extremely fast because the edges of our graph are sparse, so the number of E is nearly equal to the number of V, and V is a small value in HMER task.

\section{Experiments}
\subsection{Datasets}
We use two benchmark datasets, CROHME and HME100K, to evaluate our proposed method.

CROHME Dataset \cite{crohme14, crohme16, crohme19} is the most widely-used public dataset in the field of HMER, sourced from the Competition on Recognition of Online Handwritten Mathematical Expressions. The CROHME training set contains 8,836 HMEs within 111 symbol classes. Additionally, there are three testing sets: CROHME 2014, 2016, and 2019, comprising 986, 1,147, and 1,199 HMEs, respectively. Each HME in the CROHME dataset is stored in InkML format, capturing the trajectory coordinates of handwritten strokes. We convert this trajectory information from the InkML files into image format for both training and testing purposes.

HME100K \cite{HME100K} is a real scene HME dataset, encompassing 74,502 images for training and 24,607 images for testing. With a total of 249 symbol classes, these images originate from tens of thousands of writers and are primarily captured using cameras. As a result, HME100K exhibits a higher degree of authenticity and realism, showcasing variations in color, blur, and complex backgrounds.


\subsection{Implementation Details}
Following most previous works \cite{dwap, treedecoder, bttr, TSDNet, sam, abm, CAN-ECCV, tdv2, comer}, we keep using the same DenseNet \cite{densenet} architecture as our encoder. We use a two layer Transformer in PGD, with feed-forward dimension ratio set to 2.   And both the $\mathrm{Conv}$ dimension in VAT and Transformer dimension in PGD are set to 256. The hyper-parameter $\lambda$ in \cref{eq:loss_all} is set to 0.5 by default, and $\epsilon$ is set to 0.5 for path selection. To improve the training speed of NAMER, the estimate set $\mathbf{T}$ for VAT is obtained using a pretrained DWAP \cite{dwap} before training.

Our model is implemented on PyTorch. All experiments are conducted on four 32GB Nvidia Tesla V100 GPUs with batch size 32. The learning rate starts from 0 and monotonously increases to 2e-4 at the end of the first epoch and decays to 2e-7 following the cosine schedules, with an Adam \cite{adam} optimizer. The training epoch is set to 240 for the CROHME dataset, and 40 for the HME100K dataset. For fair comparisons, we present results with and without data augmentations. Only simple random scales and rotations are used for augmentations.

\subsection{Evaluation Metrics}
We adopt the widely used Expression recognition rate (ExpRate) as the evaluation metrics. ExpRate is defined as the percentage of correctly recognized expressions. And ExpRates under $\leq 1$ and $\leq 2$ are also used as a supplementary reference, indicating that the expression recognition rate is tolerable at most one or two symbol-level errors. Besides, GPU memory usage (MEM) and frames per sencond (FPS) are used to measure the inference costs.

\begin{table}[t]
  \caption{Performance of our model and other state-of-the-art methods on three CROHME datasets. $^*$ indicates using other informations, e.g., online stoke trajectory coordinates or extra  category of symbols. $^{\dagger}$ indicates  our reproduced result on the same training settings with NAMER. All results are reported as a percentage (\%). Our model achieves the best performance on all CROHME datasets.}
  \label{tab:SOTA}
  \centering
  \scriptsize
  \begin{tabular}{l|c|ccc|ccc|ccc}
    \toprule
    \multirow{2}{*}{Methods} & \multirow{2}{*}{Aug} & \multicolumn{3}{c|}{CROHME 2014} & \multicolumn{3}{c|}{CROHME 2016} & \multicolumn{3}{c}{CROHME 2019}\\
     & & ExpRate $\uparrow$ & $\leq 1$ $\uparrow$ & $\leq 2$ $\uparrow$ & ExpRate $\uparrow$ & $\leq 1$ $\uparrow$ & $\leq 2$ $\uparrow$ & ExpRate $\uparrow$ & $\leq 1$ $\uparrow$ & $\leq 2$ $\uparrow$ \\  
    \midrule
    UPV\cite{crohme14}  & No & 37.22 & 44.22 & 47.26 & - & - & - & - & - & - \\
    TOKYO\cite{crohme14}  & No & - & - & - & 43.94 & 50.91 & 53.70 & - & - & - \\
    WAP\cite{wap} & No & 46.55 & 61.16 & 65.21 & 44.55 & 57.10 & 61.55 & - & - & - \\
    PAL\cite{pal} & No & 39.66 & 56.80 & 65.11 & - & - & - & - & - & - \\
    TAP$^*$\cite{tap}  & No & 48.47 & 63.28 & 67.34 & 44.81 & 59.72 & 62.77 & - & - & - \\
    DWAP\cite{dwap}  & No & 50.10 & - & - & 47.50 & - & - & - & - & - \\
    DWAP-MSA\cite{dwap} & No & 52.80 & 68.10 & 72.00 & 50.10 & 63.80 & 67.40 & 47.70 & 59.50 & 63.30 \\
    DLA\cite{dla} & No & 49.85 & - & - & 47.34 & - & - & - & - & - \\
    PAL-V2\cite{palv2} & No & 48.88 & 64.50 & 69.78 & 49.61 & 64.08 & 70.27 & - & - & - \\
    WS-WAP\cite{ws-wap} & No & 53.65 & - & - & 51.96 & 64.34 & 70.10 & - & - & - \\
    TreeDecoder\cite{treedecoder} & No & 49.10 & 64.20 & 67.80 & 48.50 & 62.30 & 65.30 & 51.40 & 66.10 & 69.10 \\
    BTTR\cite{bttr} & No & 53.96 & 66.02 & 70.28 & 52.31 & 63.90 & 68.61 & 52.96 & 65.97 & 69.14 \\
    SSD\cite{ssd} & No & 53.10 & 66.00 & 71.30 & 51.70 & 64.70 & 70.10 & 53.20 & 67.60 & 73.60 \\
    TSDNet\cite{TSDNet} & No & 54.70 & 68.85 & 74.48 & 52.48 & 68.26 & 73.41 & 56.34 & 72.97 & 77.84 \\
    SAN\cite{HME100K} & No & 56.20 & 72.60 & 79.20 & 53.60 & 69.60 & 76.80 & 53.50 & 69.30 & 70.1 \\
    ABM\cite{abm} & No & 56.85 & 73.73 & 81.24 & 52.92 & 69.66 & 78.73 & 53.96 & 71.06 & 78.65 \\
    CAN-ABM\cite{CAN-ECCV} & No & 57.26 & 74.52 & 82.03 & 56.15 & 72.71 & 80.30 & 55.96 & 72.73 & 80.57 \\
    TD-V2\cite{tdv2} & No & 53.62 & - & - & 55.18 & - & - & 58.72 & - & - \\
    CoMER\cite{comer}$^{\dagger}$ & No & 58.57 & - & - & 57.89 & - & - & 59.71 & - & - \\
    SAM-CAN\cite{sam} & No & 58.01 & - & - & 56.67 & - & - & 57.96 & - & - \\
    GCN\cite{gcn}$^*$ & No & 60.00 & - & - & 58.94 & - & - & 61.63 & - & - \\
    \midrule
    $\textbf{NAMER}$ & No & \textbf{60.51} & \textbf{75.03} & \textbf{82.25} & \textbf{60.24} & \textbf{73.50} & \textbf{80.21} & $\mathbf{61.72}$ & \textbf{75.31} & \textbf{82.07} \\
    \midrule
    BTTR\cite{bttr}$^{\dagger}$ & Yes & 55.17 & 67.85 & 72.11 & 56.58 & 68.88 & 74.19 & 59.55 & 72.23 & 76.06 \\
    Ding \etal\cite{mha-stack} & Yes & 58.72 & - & - & 57.72 & 70.01 & 76.37 & 61.38 & 75.15 & 80.23 \\
  	CAN-ABM\cite{CAN-ECCV}$^{\dagger}$ & Yes & 62.47 & - & - & 59.81 & - & - & 61.72 & - & - \\
    CoMER\cite{comer} & Yes & 59.33 & 71.70 & 75.66 & 59.81 & 74.37 & 80.30 & 62.97 & 77.40 & 81.40 \\
    CoMER\cite{comer}$^{\dagger}$ & Yes & 63.35 & - & - & 60.06 & - & - & 62.21 & - & - \\
    GCN\cite{gcn}$^*$ & Yes & 64.47 & - & - & 62.51 & - & - & 64.05 & - & - \\
    \midrule
    $\textbf{NAMER}$ & Yes & \textbf{64.77} & \textbf{78.38} & \textbf{85.08} & \textbf{64.17} & \textbf{76.11} & \textbf{83.26} & \textbf{66.64} & \textbf{80.32} & \textbf{85.15} \\
       
    \bottomrule
  \end{tabular}
\end{table}
\subsection{Comparison with State-of-the-Art}
To demonstrate the superiority of our method, we compare it with previous SOTA methods. Since most of the previous methods do not use data augmentation, so we mainly focus on the results produced without data augmentation. 

\cref{tab:SOTA} displays results on the CROHME datasets. Our method attains SOTA performance across all CROHME datasets. Presently, CoMER \cite{comer} stands as the most stable SOTA HMER method. Notably, NAMER outperforms CoMER by 1.94\%, 2.35\%, and 1.49\% on CROHME 2014, 2016, and 2019, respectively, without augmentations. With augmentations, NAMER exhibits even greater superiority, surpassing CoMER by 1.42\%, 4.11\%, and 3.67\% on CROHME 2014, 2016, and 2019, respectively. These results underscore the effectiveness of the proposed method. NAMER also achieves comparable performance with the very recent published method GCN \cite{gcn}, which uses a manually defined category of symbols for supervision. Note that NAMER can be easily extended to this kind of extra supervisions, with an extra head in PGD.
\begin{table}[h]
  \caption{Results on the HME100K dataset. $^{\dagger}$ indicates our reproduced result using the same training settings with NAMER. Following previous methods' experimental settings on HME100K, no augmentation is added during training. All results are reported as a percentage (\%). NAMER achieves the best performance.}
  \label{tab:SOTA-HME100K}
  \centering
  \scriptsize
  \begin{tabular}{l|ccc}
    \toprule
    ~\multirow{2}{*}{Methods}~ & \multicolumn{3}{c}{HME100K}\\
     & ~ExpRate $\uparrow$~ & ~~$\leq 1$ $\uparrow$~~ & ~~$\leq 2$ $\uparrow$~~ \\  
    \midrule
    DWAP \cite{dwap} & 61.85 & 70.63 & 77.14 \\
    TreeDecoder \cite{treedecoder} & 62.60 & 79.05 & 85.67 \\
    BTTR \cite{bttr} & 64.10 & - & - \\
    TD-V2 \cite{tdv2}$^{\dagger}$ & 66.44 & - & - \\
    SAN \cite{HME100K} & 67.10 & - & - \\
    CAN \cite{CAN-ECCV} ]& 67.31 & 82.93 & 89.17 \\
    CoMER\cite{comer}$^{\dagger}$ & 67.90 & 79.84 & 88.03 \\
    \midrule
	\textbf{NAMER} & \textbf{68.52} & \textbf{83.10} & \textbf{89.30} \\
	\bottomrule
  \end{tabular}
\end{table}
\cref{tab:SOTA-HME100K} shows the ExpRate on the HME100K datasets. We compare our proposed method with several milestone works. Specifically, NAMER outperforms BTTR \cite{bttr}, SAN \cite{HME100K}, CAN \cite{CAN-ECCV}, CoMER\cite{comer} by 4.42\%, 1.42\%, 1.21\% and 0.62\%, respectively, showing the effectiveness in handling challenging real scene photos.

\subsection{Inference Cost}
\label{sec:speed}
To assess the efficiency of NAMER, we measured its inference costs on the CROHME dataset using a single Nvidia Tesla V100. We compared it with three representative methods: RNN-based DWAP \cite{dwap}, tree structure-based TreeDecoder \cite{treedecoder}, transformer-based CoMER \cite{comer}. For each method, we calculate the average inference time and maximum GPU memory usage across all test sets, providing the overall FPS, MEM, FLOPs, and encoder \& decoder forward times. For NAMER, the decoder forward time is the sum of VAT, PGD, and Path Selection modules. As shown in \cref{tab:speed}, NAMER achieves significant speedups: it is 13.7$\times$ and 6.7$\times$ faster than CoMER in decoding time and overall inference time, respectively. Even when compared with the simplest RNN-based DWAP baseline, NAMER still achieves a speedup of 2.36$\times$ and 1.61$\times$ in decoding time and overall FPS. Moreover, NAMER achieves significant lower memory usage, while AR models require a beam search decoding that use more memory. These improvements indicate the efficiency of the proposed non-autoregressive design.
\begin{table}[t]
  \caption{Comparison of different methods in inference costs. While keeping the same encoder, NAMER achieves a 13.7x speedup in decoding and a 6.7x speedup in overall FPS compared to the previous SOTA method CoMER. (Details of NAMER's decoding time 15.02ms: VAT=3.0ms, PGD=9.62ms, Path=2.4ms)}
  \label{tab:speed}
\scriptsize
  \centering
  \begin{tabular}{l|cc|c|c|c}
    \toprule
    \multirow{2}{*}{Methods} & \multicolumn{2}{c|}{Time (ms) $\downarrow$}  &\multirow{2}{*}{~FLOPs (G) $\downarrow$~} &\multirow{2}{*}{~FPS $\uparrow$}~ &\multirow{2}{*}{~Memory (M) $\downarrow$~} \\
	 & ~Encoder~ & ~Decoder~ & & \\
	 \midrule 
	DWAP \cite{dwap} & 18.31 & 35.41  & 30.30&18.62 & 4945 \\
	TreeDecocder \cite{treedecoder} & 18.31 & 115.81  & 55.39 &7.46 & 6290 \\
	CoMER \cite{comer} & 18.31 & 206.12  & 174.33 &4.46 & 13441 \\
	\midrule 
	\textbf{NAMER} & 18.31 & \textbf{15.02}  & \textbf{11.39} &\textbf{30.00} & \textbf{2569} \\
	\bottomrule
  \end{tabular}
\end{table}

\subsection{Ablation Study}
\label{sec:ablation}
\textbf{Training targets of VAT}. VAT is trained under the bipartite matching results between a pretrained HMER model $\mathrm{M}^*$ and the predicted visible probability $\mathbf{P}$. Here we carefully explore the influence of different pretrained HMER models and different bipartite matching area for each token. As shown in \cref{tab:ablation-vat}, using different methods for generating bipartite matching targets is not sensitive for our model training, the performance is almost the same. According to \cref{tab:SOTA}, performance of these for methods range across \textasciitilde10\% of ExpRate, proving that the key goal of VAT is to instantiate visible tokens and the approximate location is enough for HMER. In fact, bipartite matching training allows NAMER to learn optimal solutions within a tolerance window, enabling VAT to robustly learn from various coarse targets. Consequently, methods with lower ExpRate may outperform CoMER as a VAT target slightly.

 In \cref{tab:vat-size}, we investigate the effective matching area for each token during bipartite matching, varying it from 3 to 7. The results indicate that a kernel of $5\times5$ is the best for bipartite matching, it is make sense because a character symbol is usually \textasciitilde40 pixel. (Remind that bipartite matching is performed at the feature map of $\frac{H}{8}\times\frac{W}{8}$)
\begin{table}[t!]
\scriptsize
\caption{Ablation study on different bipartite matching targets of VAT. All results are obtained on CROHME 2014 using the proposed NAMER without augmentations.}
  \label{tab:ablation-vat}
  \centering
  \scriptsize
  \begin{tabular}{c|cccc}
    \toprule
	VAT Target & ~DWAP \cite{dwap}~ & ~BTTR \cite{bttr}~ & ~CAN \cite{CAN-ECCV}~  & ~CoMER \cite{comer}~ \\
	\midrule 
	ExpRate & 60.51 & \textbf{60.65} & 60.55 & 60.35 \\
	\bottomrule
  \end{tabular}
\end{table}
 \begin{table}[t!]
\scriptsize
  \caption{Ablation study on kernel sizes for computing VAT's matching cost matrix.}
  \label{tab:vat-size}
  \centering
  \begin{tabular}{c|ccc}
    \toprule
	~$k_{m} \times k_{m}$~ & ~3$\times$ 3~ & ~5$\times$ 5~ & ~7$\times$ 7~\\
	\midrule 
	ExpRate & 59.74 & \textbf{60.51} & 60.35 \\
	\bottomrule
  \end{tabular}
\end{table}
\begin{table}[t!]
\centering
  \caption{Ablation study on different heads of PGD. All results are obtained on CROHME dataset without data augmentations.} 
  \label{tab:ablation-pgd}
\scriptsize
   \begin{tabular}{ccc|ccc}
    \toprule
    ~\multirow{2}{*}{LCH}~ & ~\multirow{2}{*}{RCH}~ & ~\multirow{2}{*}{SCH}~ & \multicolumn{3}{c}{ExpRate $\uparrow$} \\
     & &  & ~2014~ & ~2016~ & ~2019~\\ 
    \midrule 
   \checkmark & & & 55.03 & 54.84 & 57.13 \\
    & \checkmark & & 54.62 & 55.01 & 56.38 \\
   \checkmark & \checkmark & &  55.94 & 56.32 & 58.38 \\	
   \checkmark & & \checkmark & 58.88 & 58.15 & 60.22 \\ 
    & \checkmark & \checkmark & 57.26 & 57.98 & 58.97 \\ 
   \checkmark & \checkmark & \checkmark & \textbf{60.51} & \textbf{60.24} & \textbf{61.72} \\	
	
	\bottomrule
  \end{tabular}
\end{table}

\textbf{Effectiveness of modules in PGD}. To study the contributions of each head in PGD, we conduct detail ablation experiments on \cref{tab:ablation-pgd}, where we alternately control the use of three heads: LCH, RCH, and SCH. The experimental results illustrate three key conclusions:
\begin{enumerate}
  \item With only one head LCH or RCH, it can achieve comparable baselines, proving the effectiveness of both the VAT and connectivity prediction heads;
  \item The merging of left and right connectivity scores can help enhance the accuracy of the graph structure (\raisebox{-0.5ex}{\textasciitilde}2.5\% with SCH and \raisebox{-0.5ex}{\textasciitilde}1\% without SCH);
  \item Based on VAT's language feature, position feature, and image feature, SCH can indeed do a vision-language modeling for correcting misclassified tokens in VAT, upgrading the ExpRate by \raisebox{-0.5ex}{\textasciitilde}3.5\% in average.
\end{enumerate}

\subsection{Case Study}
\label{sec:case}
\begin{figure}[t]
  \centering
	\includegraphics[width=0.99\linewidth]{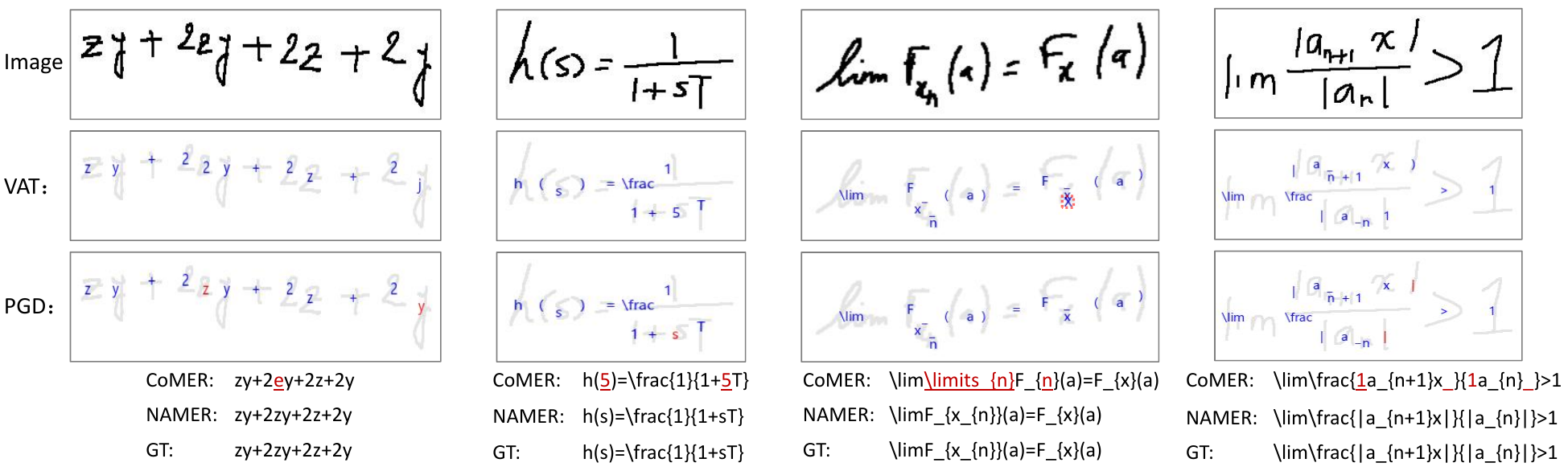}
  \caption{Visualization and comparison of NAMER results at different stages: the first three image rows display visualizations from NAMER, while the last three text rows present comparisons with CoMER. Characters in red color on line PGD denote that PGD has made a correction on VAT's results.}
  \label{fig:case_study}
\end{figure}

We select four typical examples in \cref{fig:case_study} to assess NAMER's capabilities and make comparsions. The first two images exhibit ambiguities between ``2''-``z'', ``y''-``j'', and ``5''-``s''. In NAMER, VAT makes initial mistakes on these symbols, yet the PGD effectively rectifies them by leveraging comprehensive visual and linguistic contexts. Conversely, CoMER tends to misclassify these ambiguous symbols. In the third case, VAT misidentifies an additional symbol ``X'', which PGD successfully eliminates as a redundant token. Notably, NAMER demonstrates a more consistent ability to recognize local relations, correctly identifying the two adjacent ``\_'' symbols, whereas CoMER fails to do so accurately. The final case highlights the error accumulation issue observed in AR-based methods; CoMER misidentifies ``$|$'' as ``1'' and subsequently fails to recognize the subsequent ``$|$'' tokens correctly. In contrast, NAMER exhibits superior robustness and capability in handling such situations. Additionally, we have observed that when PGD rectifies symbols, the cross and self-attention maps in \cref{qkv} usually focus on the corresponding image context areas and the correct VAT tokens. This demonstrates the capability and interpretability of PGD.

\subsection{Limitation}
\label{sec:limitation}
Due to the incapability of PGD to insert new tokens, limitations of NAMER lies in handling extreme cases where VAT fails to recall relational tokens (e.g., `\textasciicircum', `\textunderscore'). These cases are rare, typically arising from serious handwriting errors that necessitate human understanding of context for correction (e.g., writing x\textasciicircum2+y\textasciicircum2=R\textasciicircum2 as x\textasciicircum2+y2=R\textasciicircum2). Solving such cases could be a future work, specifically by empowering PGD with insertion capability. 

\section{Conclusion}
This paper presents NAMER, a novel NAR Modeling paradigm designed for HMER. Unlike prevailing AR approaches, NAMER adopts a bottom-up, coarse-to-fine methodology. NAMER firstly predicts visible symbols and local relations, then refines them and establishes connectivities in a NAR manner. To achieve this, a Visual Aware Tokenizer, a Parallel Graph Decoder, and a Bipartite Matching Training strategy are elaborately designed. To the best of our knowledge, NAMER is the first work to explore NAR in HMER, leveraging comprehensive visual and linguistic contexts. Experiments have demonstrated the effectiveness and efficiency of our method. In future work, we intend to extend the capabilities of NAMER to tackle more intricate structural recognition tasks, such as interpreting chemical structure images and complex CAD images.

\bibliographystyle{splncs04}
\bibliography{paper7448_eccv2024_camera_ready}

\title{Supplementary Materials of \\ NAMER: Non-Autoregressive Modeling for Handwritten Mathematical Expression Recognition} 

\titlerunning{Supplementary Materials of NAMER}

\author{Chenyu Liu\inst{1,2} \and
Jia Pan\inst{2} \and
Jinshui Hu\inst{2} \and
Baocai Yin\inst{2} \and 
Bing Yin\inst{2}\and \\ 
Mingjun Chen\inst{2}\and 
Cong Liu\inst{2}\and 
Jun Du\inst{1}$^\dag$ \and 
Qingfeng Liu\inst{1,2}
}

\authorrunning{C. Liu et al.}

\institute{University of Science and Technology of China, Hefei, China \and
iFLYTEK Research, Hefei, China \\
\email{cyliu7@mail.ustc.edu.cn, jundu@ustc.edu.cn}\\
}

\maketitle
\renewcommand{\thefootnote}{}
\footnotetext[2]{$^\dag$ Corresponding author.}

\section{NAMER Label Token Details}
\label{sec:intro}

In this section, we provide details of labels in NAMER, including all local relation tokens and the attached imaginary tokens for them.

\begin{table}[ht]
   \caption{Local Relation Tokens in VAT: Our defined imaginary relationship symbols are common to both the CROHME \cite{crohme14, crohme16, crohme19} dataset and the HME100K \cite{HME100K} dataset. Due to HME100K being a more complex dataset with a larger set of symbol classes, it contains a larger number of handwritten structural elements compared to CROHME.}
  \centering
  \begin{tabular}{l|c|c}
    \toprule
     & Dataset & Label Symbol \\
    \midrule
	\multirow{3}{*}{~~~IRS~~~} & \multirow{3}{*}{All} &  \textasciicircum\\
	 & & \textunderscore \\
	 & & \textbackslash limits \\
	 \midrule
	 \multirow{2}{*}{~~~HSE~~~} & \multirow{2}{*}{~~~CROHME \cite{crohme14, crohme16, crohme19}}~~~ & \textbackslash frac \\
	  &  & \textbackslash sqrt \\
	 \midrule
	 \multirow{11}{*}{~~~HSE~~~} & \multirow{11}{*}{HME100K \cite{HME100K}} & \textbackslash frac \\
	  &  & \textbackslash sqrt \\  
	  &  & \textbackslash dot \\
	  &  & \textbackslash ddot \\
	  &  & \textbackslash boxed \\
	  &  & \textbackslash widehat \\
	  &  & \textbackslash overline \\
	  &  & \textbackslash xlongequal \\
	  &  & \textbackslash textcircled \\
	  &  & \textbackslash xrightarrow\\
	  &  & ~~~\textbackslash overrightarrow~~~ \\	 
	\bottomrule
  \end{tabular}
  \label{tab:vat}
\end{table}

\subsection{Local Relation Tokens in VAT}

As mentioned in Section 3.2, local relation tokens in the Visual Aware Tokenizer (VAT) consists of Handwritten Structural Elements (HSE) and Imaginary Relationship Symbols (IRS). Details of HSE and IRS in different datasets are shown in \cref{tab:vat}.

\subsection{Attached Imaginary Tokens in PGD}
As outlined in Section 3.3, after obtaining all visible tokens $\mathbf{y}_{\mathrm{vat}}$ from VAT, the respective attached imaginary tokens related to the local relation tokens in $\mathbf{y}_{\mathrm{vat}}$ are added. For instance, in a mathematical expression $x^{yz}+1$ with VAT results \{``x'', ``\textasciicircum'', ``y'', ``z'', ``+'', ``1'', \}, the ending token ``\}'' related to ``\textasciicircum'' must be added to fully recover the upper corner relation structure.

The definition of an attached imaginary token is straightforward: it represents the ending token for the local relation token, as illustrated in \cref{tab:pgd}. It's important to note that most local relation tokens have a single ``\}'' ending imaginary token, whereas ``\textbackslash frac'' and ``\textbackslash sqrt'' have two ending imaginary tokens. We have defined all attached tokens as ``\}'', ensuring uniformity with our label tokens in LaTeX.

Note that all the attached imaginary tokens are automatically generated by scanning all label strings using a simple Python script, not manually maintained. This script is written based on LaTeX grammar of special tokens. 

\begin{table}[ht]
  \caption{Details of all attached imaginary tokens. It is defined as the ending tokens for each local relation token.}
  \centering
  \begin{tabular}{|c|c|}
    \hline
     ~~~Local Relation Token~~~ & ~~~Attached Imaginary Token~~~ \\
     \hline
	 \textasciicircum & one '\}' for ending \\
	 \hline
	 \textunderscore & one '\}' for ending \\
	 \hline
	 \textbackslash limits & one '\}' for ending \\
	 \hline
	  \textbackslash dot &one '\}' for ending\\
	  \hline
	  \textbackslash ddot &one '\}' for ending\\
	  \hline
	  \textbackslash boxed &one '\}' for ending\\
	  \hline
	  \textbackslash widehat &one '\}' for ending\\
	  \hline
	  \textbackslash overline &one '\}' for ending\\
	  \hline
	  \textbackslash xlongequal &one '\}' for ending\\
	  \hline
	  \textbackslash textcircled &one '\}' for ending\\
	  \hline
	  \textbackslash xrightarrow &one '\}' for ending\\
	  \hline
	  \textbackslash overrightarrow &one '\}' for ending\\	
	  \hline
	\multirow{2}{*}{\textbackslash frac} & one '\}' for numerator ending, \\
	 & one '\}' for denominator ending \\
	\hline
	  \multirow{2}{*}{\textbackslash sqrt} &one '\}' for radicand ending,\\
	  & one '\}' for index ending \\
 
	\hline
  \end{tabular}

  \label{tab:pgd}
\end{table}

\begin{figure*}[ht!]
\centering
\begin{minipage}{0.95\textwidth}
\begin{lstlisting}[style=PythonStyle, caption=Example Python Code for Dynamic Assignment Approach in VAT, label=python_code]
# Python code for dynamic assignment in VAT. It's an example for one training sample.
# And it's easy to extend this code to support batch training.
import torch
from scipy.optimize import linear_sum_assignment

def vat_matching(P_vat, y_label, T_dwap, vocab_size, assign_kernel=(5, 5)):
    '''
    Parameters
    ----------------
    P_vat: FloatTensor, [K+1, H/8, W/8]
        predicted probabilities of VAT, which is the Eqn (2)'s \mathbf{P}
    y_label: LongTensor, [L, 1]
        all visible tokens, y_1, y_2, ..., y_L
    T_dwap: LongTensor, [L, 2]
        estimated spatial positions of y_label using a pretrained DWAP
    vocab_size: int
        vocabulary size
    assign_kernel: tuple (int, int)
        effective local matching window k_m * k_m for P_vat and T_vat
        
    Returns
    ----------------
    P_target: LongTensor, [H/8, W/8]
        The bipartite matching result of training target \mathbf{P}^* for VAT
    '''
    _, H, W = P_vat.shape  # get sizes of feature map
    L, _ = y_label.shape      # get length of visible tokens
    
    assign_pad = (assign_kernel[0]//2, assign_kernel[1]//2)  # compute paddings
    y_indices = torch.arange(L).cuda()  # a tensor for indexing

    # convert T_vat to a 3D matrix
    T_vat_mat = torch.zeros([L, H, W]).cuda()
    T_vat_mat[y_indices, T_dwap[y_indices, 0], T_dwap[y_indices, 1]] = 1.0

    # set effective matching positions using maxpooling
    T_vat_mat = torch.nn.functional.max_pool2d(T_vat_mat.unsqueeze(0), kernel_size=assign_kernel, stride=(1,1), padding=assign_pad)[0]

    # computing overall distance matrix for assigning H*W predictions to T tokens
    dist_mat = torch.zeros([L, H, W]).cuda()
    dist_mat[y_indices] = (P_vat[y_label[y_indices]] - T_vat_mat[y_indices]).abs()
    # limiting matching within a k_m x k_m window for each token
    dist_mat = dist_mat * T_vat_mat + (1 - T_vat_mat) * 1e6

    # hungarian algorithm for optimal assignment
    row_ind, col_ind = linear_sum_assignment(dist_mat.view(L, -1).cpu().numpy())
    h_ind = col_ind // W
    w_ind = col_ind % W

    # generating P_target for training VAT module
    P_target = (torch.ones([H, W]) * vocab_size).long() # initialized as all \varnothing.
    P_target[h_ind, w_ind] = y_label[row_ind]
    return P_target
\end{lstlisting}
\end{minipage}
\end{figure*}

\section{Details of Dynamic Assignment Process}
\label{sec:formatting}

In this section, we present detailed code for the proposed bipartite matching-based dynamic assignment approach, which is initially introduced in Section 3.2 to generate training targets for VAT.

The code snippet in \cref{python_code} outlines the matching and assignment process. Specifically, given the estimated spatial positions $\mathbf{T} \in \mathbb{Z}^{L \times 2}$, for the parallel matching of all batches and tokens, we firstly convert $\mathbf{T}$ to a 3D matrix resembling an indicator function (line 34 in \cref{python_code}). Subsequently, a max-pooling operation is applied to achieve local $k_m \times k_m$ window-based matching (line 37). Next, the overall distance matrix is computed between VAT's predicted probabilities $\mathbf{P}$ and the converted $\mathbf{T}_{mat}$. Finally, the Hungarian algorithm is employed for optimal assignment, resulting in the training target $\mathbf{P}^*$ for VAT.

\section{Ablation Study on Path Selection}
\label{sec:path}
Path selection involves directed acyclic graph (DAG) construction and the longest path selection. We performed ablations on the edge $\mathrm{E}_{ij}$ of Eq. (10), finding a balanced 1:1 ratio of left-to-right (L2R) and right-to-left (R2L) score yields the optimal results, as shown in \cref{tab:path}.

 \begin{table}[h!]
  \caption{Ablation study on edge value $\mathrm{E}_{ij}$ of DAG construction for Path Selection.}
  \label{tab:path}
  \centering
  \begin{tabular}{c|ccccc}
    \toprule
	~~~L2R:R2L~~~ & ~~~1:0~~~ & ~~~1:0.5~~~ & ~~~1:1~~~ & ~~~0.5:1~~~ & ~~~0:1~~~\\
	\midrule 
	ExpRate & 58.88 & 59.43 & \textbf{60.51} & 59.13 & 57.26 \\
	\bottomrule
  \end{tabular}
\end{table}

%
%
\bibliographystyle{splncs04}
\end{document}